\documentclass[letterpaper, 10 pt, conference]{ieeeconf}  

\IEEEoverridecommandlockouts                              

\usepackage{graphics} 
\usepackage{epsfig} 
\usepackage{amsmath} 
\usepackage{amssymb}  
\usepackage{bm}
\usepackage[ruled]{algorithm2e}
\usepackage{xcolor}

\title{\LARGE \bf
State Estimation and Environment Recognition for Articulated Structures via Proximity Sensors Distributed over the Whole Body
}

\author{Kengo Iwao$^{1}$, Hikaru Arita$^{1}$ and Kenji Tahara$^{1}$
\thanks{$^{1}$The authors are with Department of Mechanical Engineering,
Kyushu University, Fukuoka 819-0395, JAPAN.
The corresponding author is Hikaru Arita.
        {\tt\small iwao@hcr.mech.kyushu-u.ac.jp;[arita,
tahara]@ieee.org}}%
}

\begin{document}

\maketitle
\thispagestyle{empty}
\pagestyle{empty}

\begin{abstract}

For robots with low rigidity, determining the robot's state based solely on kinematics is challenging. 
This is particularly crucial for a robot whose entire body is in contact with the environment, as accurate state estimation is essential for environmental interaction. 
We propose a method for simultaneous articulated robot posture estimation and environmental mapping by integrating data from proximity sensors distributed over the whole body.
Our method extends the discrete-time model, typically used for state estimation, to the spatial direction of the articulated structure. 
The simulations demonstrate that this approach significantly reduces estimation errors.
\end{abstract}

\section{INTRODUCTION}

The posture of an articulated robot can generally be determined from the joint angles via kinematics. 
This is true for robots in which each link is rigid and each joint angle measurement is accurate, such as industrial manipulators.
However, not all modern robots have such characteristics.
For example, the numbers of robots that can perform detailed tasks by learning with inexpensive hardware \cite{gello, mobilealoha, lowcost-learning} and lightweight arms designed to be mounted on mobile robots \cite{mobilealoha, Xarm} have increased in recent years.
The lightweight and inexpensive features of such robots 
mean that the rigidity of each link and the accuracy of joint angle measurements tend to be lower than in previous robots.
When the deformation of these less rigid links and the angular errors of the joints are considered, kinematics alone cannot accurately estimate the posture, which can be an important problem in situations that require detailed work. 

Research has been done on methods of ensuring accurate end-effector positioning, including end-effector position correction through marker observation \cite{endef1, endef2} and arm tracking through depth images from cameras mounted separately from the joints \cite{tracking}.
However, there are situations in which tracking an end-effector is not sufficient.
A robot that moves within the environment and has an articulated structure, such as in \cite{mobilealoha, Xarm}, needs information of external information regarding its entire body and whole-body state to the environment because the entire body may come into contact with the environment.
These issues are equally applicable to snake robots.
Snake robots navigate in unknown environments by maintaining full-body contact, making it crucial for them to ascertain their own postures relative to the environment.
Furthermore, when traversing uneven terrain, these robots often lift parts of their bodies while maintaining contact with the environment. 
Consequently, lightweight links are frequently employed, many of which are prone to deformation.
For such robots, where task execution based solely on internal information such as kinematic models is challenging, it is necessary to acquire information on the external environment and simultaneously estimate the overall state of the robot with respect to the external environment.
One approach is to use simultaneous localization and mapping (SLAM) for this purpose.

The SLAM method is generally used for mobile robots, but several studies have applied SLAM to robot arms to address the uncertainty of the state of a robot due to factors such as gear backlash and nonrigid deformation.
For example, ARM-SLAM \cite{arm-slam} uses a depth camera attached to the end-effector to perform SLAM, which reduces uncertainty and simultaneously yields information on the external environment.
In addition, a method has been proposed to attach RGB-D cameras to multiple joints of a soft robot and perform SLAM to estimate the robot's configuration \cite{arm-slam2}.
While these studies have successfully reduced angular errors in the arms, these cameras cannot obtain information at close range because their field of view is completely obstructed when the camera is too close to the environment, and is not suitable for the situations involving contacts.
Morever, our goal is to acquire information on the environment surrounding the whole body, and it is difficult to achieve this with a camera that has a limited focus, angle of view, and position of placement.

As proximity perception information for motions involving contact, proximity sensor data are utilized as visual information \cite{survey1}.
Because these sensors are small and lightweight, they can be placed on the robot's entire body to acquire the environmental information surrounding the entire body.
There is increasing research on attaching proximity sensors to link surfaces and using external information from the robot surface.
For example, information obtained from proximity sensors that cover the surface of a link has been used to perform collision avoidance for a robot \cite{proximity, proximity2, proximity3}.
Some studies have also used proximity sensors attached to each joint of a snake robot to detect planes in the area of the entire body \cite{snakeprox}.
Utilizing the idea of these research, we focus on obtaining observation data for whole-body SLAM from optical proximity sensors distributed across the entire body of the robot.

By distributing sensors over the whole body, each link can have its own external information, allowing SLAM to be performed individually for each link.
Using this feature, we propose extending the discrete-time model of SLAM in the spatial direction by recursive estimation of the whole body.
In general, SLAM uses the idea of the Bayesian filter, which discretizes the continuous motion of the robot, for estimating the robot's motion.
We focus on applying this discrete model to articulated robot structures.
By recursively describing the state of each articulated link at the same time, state variables and their uncertainties can be propagated along the spatial direction.
This enables the cumulative errors that occur with each successive link to be reduced.

In summary, the statement of our problem and the corresponding proposal to solve it are as follows:

\begin{enumerate}
\setlength{\leftmargini}{-1cm}
\item[-]Problem Statement
\begin{itemize}
\item For articulated robots and soft robots that are constructed with nonrigid components for the purpose of weight reduction and simplification, it is difficult to determine their states solely through kinematics.
\item Moreover, as these robots are often utilized in complex environments where full-body contact with their surroundings is likely, it is essential to determine the posture of the whole body relative to the environment.
\end{itemize}
\item[-]Proposed Approach
\begin{itemize}
\item To address these issues, we propose a method for estimating a robot's state relative to its environment by distributing proximity sensors across the entire body of the robot and performing SLAM on observations from the full body.
\end{itemize}
\item[-]Key Innovation
\begin{itemize}
\item We reduce the accumulation of errors by extending the structure of the discrete-time model used in SLAM to the spatial direction along the links.
\end{itemize}
\end{enumerate}

We first explain our proposed method in Section \ref{Proposed Method}.
Three simulations for validating the proposed method are described in Section \ref{Simulation}.
Finally, the advantages of the proposed method obtained from the simulations, as well as its applicability, are discussed in Section \ref{Discussion}, and Section \ref{Conclusion} concludes this paper.

\section{Proposed method}\label{Proposed Method}
\subsection{Mathematical Notation}
To describe the estimation methods, this paper uses several symbols, as shown in Table \ref{description}.
Furthermore, we define a single full-body estimation process at a given time as a ``step".
\begin{table}[tb]
\begin{center}
\caption{Symbol Description}
\begin{tabular}{l|p{7cm}}
\hline
$\mathbf{x}_{i,k}$ & state of the $i$th link at the $k$th step. $i=0$ is the state of the root.\\ 
$\mathbf{\bar{x}}$ & final estimated state\\
$\mathbf{\tilde{x}}$& error state with respect to the true state\\
$\mathbf{\hat{x}}^\kappa$ & state obtained in the $\kappa$th iterated Kalman filter. $\mathbf{\hat{x}}^{0}$ prior estimated state.\\
$\mathbf{\tilde{x}}^\kappa$ & error state between $\mathbf{\hat{x}}^\kappa$ and $\mathbf{\hat{x}}^{\kappa+1}$\\
$\mathbf{\tilde{x}}^{0,\kappa}$ & error state of $\mathbf{\hat{x}}^{0}$ with respect to $\mathbf{\hat{x}}^\kappa$.\\
$\mathbf{p}_j$ & $j$th point measurement from the sensor for a single estimation. \\
$\mathbf{q}_j$ & point on the map corresponding to $\mathbf{p}_j$\\
$L$, $W$ & link frame and world frame\\
\hline
\end{tabular}
\label{description}
\end{center}
\end{table}

\subsection{Problem Statement}
\begin{figure}[tb]
\begin{center}
\includegraphics[width=\columnwidth]{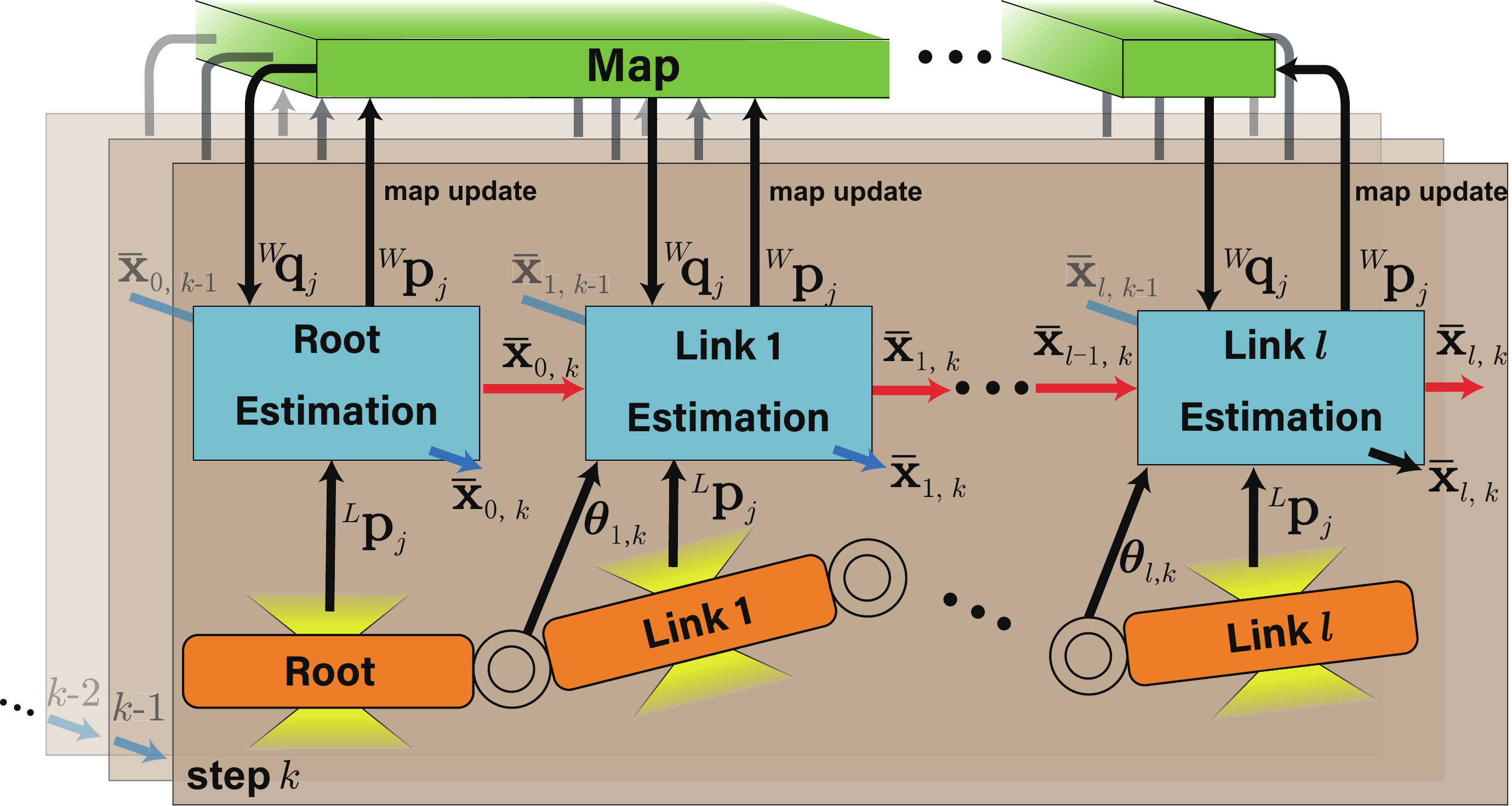}
\caption{Overview of the method. $\theta$ is the angle obtained from the encoder. The red arrows represent the spatial direction, and the blue arrows represent the temporal direction.}
\label{system-overview}
\end{center}
\end{figure}

For simplicity, a common and straightforward model of a articulated structure is considered in this paper, as illustrated in Fig.~\ref{system-overview}.
The root is the reference link of an articulated structure, such as the base of a manipulator or the head of a snake-like robot.
The rotational axis of each joint can be defined arbitrarily, allowing the entire structure to perform three-dimensional motion.
The origin of each link frame is defined as the center of the joint on the root side where the link is connected.
The joint angle can be obtained from the encoders at each joint, which includes bias, and external information data can be obtained from the proximity sensors covering each link, which includes white noise.

To consider a feasible method, we assume a model that uses a VL53L5C \cite{STMicro} as an example of an existing ToF-type proximity sensor that can cover a link. 
The VL53L5C is small (6.4 mm $\times$ 3.0 mm $\times$ 1.5 mm) and can be distributed on links. 
We assume that these sensors are placed 
along the circumference 
of the link.
The sensor uses multiple light beams emitted from a single unit to acquire distance data from target objects.
While the sensor has relatively low responsiveness compared with other proximity sensors, a single unit is capable of obtaining environmental information at a frequency comparable to those of the LiDAR systems commonly employed in SLAM applications.
This configuration enables the comprehensive environmental data surrounding the entire body to be treated as point cloud data derived from proximity sensors.

\subsection{Foundational method}
Point cloud-based SLAM methods commonly extract features, such as planes and edges, from sensor point cloud data to reduce the number of computations \cite{loam}.
However, the method of feature extraction changes according to the method of gathering points \cite{fastlio2}.
In our assumed situation, where many sensors are installed on the link surface, the sensor model must be constructed every time the sensor arrangement changes.
To avoid this, we applied the technique of FAST-LIO2 \cite{fastlio2} to our state estimation method.
One of the advantages of FAST-LIO2 is its ability to use raw point cloud data for estimation while maintaining a low computational load; it can do this because it uses a point cloud management method involving {\it ikd-Tree} \cite{fastlio2} and an error-state iterated Kalman filter, which has a computational load that depends on the state dimension instead of the measurement dimension \cite{fastlio}.

We developed a method for whole-body SLAM of articulated structures by applying FAST-LIO2 to our two key ideas: the acquisition of external environment information by proximity sensors and the information propagation in the spatial direction using the articulated link structure.

\subsection{System Overview}\label{system overview}
The system overview is shown in Fig.~\ref{system-overview}.
The proposed method is divided into two stages at each step: the estimation of the root state $\mathbf{x}_{0,k}$ and the subsequent estimation of the state $\mathbf{x}_{i,k}\, (i>0)$ of each link.
The root estimation stage at each step follows the same process as in conventional SLAM via a discrete-time model.
By combining the prediction based on the root state $\mathbf{\bar{x}}_{0,k-1}$ obtained from the previous step with the current observations $\mathbf{p}_j$, we can perform state estimation of the root and acquire information on the environment surrounding the root.

Once the estimation of the root state is completed, for the subsequent links, we can construct a model that utilizes the constraint that all links are connected at the same time.
In other words, after the root estimation, 
we make a prediction by recursively describing the state of each link along the link direction and then combine this prediction 
with observations to perform estimation.
In this process, estimation proceeds from the root to the end link in the same time series, which we call the spatial direction.
However, not all state variables can be propagated spatially. 
Some state variables, such as the biases of the joints, change in unique ways for each joint, regardless of the link constraints. 
Therefore, as with the root, these variables are predicted on the basis of their past states.
In this process, estimation proceeds in the temporal direction.
Thus, the states of the links after the root are predicted by combining the state variables propagated spatially and those propagated temporally. 
These predictions are then integrated with observations to estimate the state relative to the environment and acquire environmental information. 

Once the estimation of all the links, i.e., the estimation of the full-body posture, is complete, the estimation for the next step begins again from the root.
The point cloud data obtained from the proximity sensors are converted to the world coordinate system after the state of each link is estimated and are added to the map as information by {\it on-tree downsampling}\cite{fastlio2}.
This allows the amount of available map information to increase with each successive estimation of the link, even within the same step.

Integration with observations is performed via an iterated extended Kalman filter.
The details of this process, including the model, are described in Sections \ref{articulated structure model}$\sim$\ref{iterated extended kalmam filter}.

\subsection{Articulated Structure Model}\label{articulated structure model}
The root state $\mathbf{x}_{0,k}$ includes the position $\mathbf{p}_{0,k}\in\mathbb{R}^3$ and orientation $\mathbf{R}_{0,k}\in \mathrm{SO(3)}$.
The state of the root at the $k$th step can be expressed as follows.
\begin{align}
\mathbf{p}_{0,k} &= \mathbf{p}_{0,k-1} + \boldsymbol{\omega}_{\mathbf{p}_k} \label{root pos model}\\ 
\mathbf{R}_{0,k} &= \mathbf{R}_{0,k-1}\mathrm{Exp}(\boldsymbol{\omega}_{\mathbf{R}_k})\label{root rot model}
\end{align}
In this work, temporal changes in position and orientation with respect to the previous step are modeled by Gaussian noise as a random walk process, where $\boldsymbol{\omega}_{\mathbf{p}_k}\in\mathbb{R}^3$ is the amount of change in position; $\boldsymbol{\omega}_{\mathbf{R}_k}\in\mathbb{R}^3$ is a vector of rotation axes.
 In the case of snake robots or mobile manipulators, the root often includes IMUs or wheel odometry. 
It is therefore possible to construct a root model that incorporates the available odometry.
$\mathrm{Exp}(\mathbf{n})\in \mathrm{SO(3)}$ is the matrix exponential expressed by the {\it Rodrigues rotation formula} as follows:
\begin{align}
\label{exp}
\mathrm{Exp}(\mathbf{n})= \mathbf{I} + \sin \|\mathbf{n}\|\left[\frac{\mathbf{n}}{\|\mathbf{n}\|}\right]_\times+ (1-\cos\|\mathbf{n}\|)\left[\frac{\mathbf{n}}{{\|\mathbf{n}\|}}\right]^2_\times
\end{align}
where $\mathbf{I}$ represents the identity matrix and where $\left[\cdot\right]_\times$ denotes an operator that transforms an $\mathbb{R}^3$ vector to a skew-symmetric $\mathbb{R}^{3\times3}$ matrix.

The state of the link following the root $\mathbf{x}_{i,k} \, (i>0)$ includes not only $\mathbf{p}_{i,k}$ and $\mathbf{R}_{i,k}$ but also the angular bias $b_{i,k}\in\mathbb{R}$ of the joint.
The state of the $i$th link at the $k$th step can be described as follows.
\begin{align}
{b}_{i,k} &=
{b}_{i,k-1} + {\omega}_{{b}_{i,k}}\\
\mathbf{p}_{i,k} &=
\mathbf{p}_{i-1,k} + \mathbf{R}_{i-1,k}{}^{i-1}\!\mathbf{p}_i\\
\mathbf{R}_{i,k} &= 
\mathbf{R}_{i-1,k}\mathrm{Exp}\left\{\boldsymbol{\theta}_{i,k}-\left({b}_{i,k-1}-\omega_{\theta_{i,k}}\right)\frac{\boldsymbol{\theta}_{i,k}}{\|\boldsymbol{\theta}_{i,k}\|}\right\}\label{model 5}
\end{align}
The position $\mathbf{p}_{i,k}$ and orientation $\mathbf{R}_{i,k}$ are determined by the state of the $i-1$th link propagated in the spatial direction from the same step, while the bias ${b}_{i,k}$ is determined by the state of the $i$th estimated in the past step.
The amount of change in the bias at each step is modeled by the Gaussian noise ${\omega}_{{b}_{i,k}}$
as a random walk process.
The relative position vector ${}^{i-1}\!\mathbf{p}_{i}\in\mathbb{R}^3$ between the links is determined by the shape of each link.
The change in posture between links is represented by a rotation axis vector $\boldsymbol{\theta}_{i,k}$ derived from the joint angles considering the bias $b_{i,k-1}$ and measurement noise ${\omega}_{{\theta}_{i,k}}$.

\subsection{Measurement Model}
The sensors on each link acquire a point cloud $  \{\mathbf{p}_j \mid j = 1,2,\ldots,m\}$ once per step.
The sensor model employed in this study is fundamentally similar to that described in \cite{fastlio2}.
For a detailed derivation of this sensor model, readers are directed to \cite{fastlio2}.

The acquired point cloud is converted from a link frame to a global frame according to the predicted position and orientation of each sensor and then projected onto the map.
Assuming that the projected points should be in the local plane on the map, as in \cite{fastlio2}, the implicit sensor model is constructed as follows:
\begin{align}
\mathbf{n}_j\left[\mathbf{R}_{i,k}\left\{\mathbf{R}_{S_i}^{L_i}\left(\mathbf{p}_j -\boldsymbol{\upsilon}_j\right)+\mathbf{p}_{S_i}^{L_i}\right\}+\mathbf{p}_{i,k}-\mathbf{q}_j\right]=0\label{measurement model2}
\end{align}
where $\mathbf{n}_j$ is the normal vector of the local plane formed by the neighborhood points of the map that include $\mathbf{q}_j$ when the points $\mathbf{p}_j$ are projected onto the map; $\mathbf{p}_{S_i}^{L_i}$ and $\mathbf{R}_{L_i}^{S_i}$ are the position vector and orientation matrix of the sensor in the $i$th link frame, respectively; and $\boldsymbol{\upsilon}_j$ is the measurement noise. 

\subsection{Iterated Extended Kalman Filter}\label{iterated extended kalmam filter}
The state estimation is performed via an iterated extended Kalman filter using an error state model as in \cite{fastlio, iterated}.
The use of error states allows all state quantities, including attitudes, to be expressed in $\mathbb{R}^3$, which is a minimum representation \cite{error_state}.
The error state for each state is defined as follows:
\begin{align}
\mathbf{\tilde{a}}&=\mathbf{a}-\mathbf{\bar{a}}&\mathbf{a}\in\mathbb{R}^3 \label{error state vec}\\
\mathbf{\tilde{A}}&=\mathrm{Log}\left(\mathbf{\bar{A}}^\top\mathbf{A}\right)&\mathbf{A}\in \mathrm{SO(3)} \label{error state so3}
\end{align}
where $\mathrm{Log}\left(\cdot\right)\in\mathbb{R}^3$ is the inverse function of (\ref{exp}); its normalized vector represents the rotation axis of the posture error, and its magnitude represents the rotation angle of the posture error.
With the introduction of error states, each state can be represented by a single vector as follows ($i>0$):
\begin{align}
\mathbf{\tilde{x}}_{0,k}&= \begin{bmatrix}\mathbf{\tilde{p}}_{0,k}^\top&\mathbf{\tilde{R}}_{0,k}^\top\end{bmatrix}^\top\\
\mathbf{\tilde{x}}_{i,k}&= \begin{bmatrix}\mathbf{\tilde{p}}_{i,k}^\top&\mathbf{\tilde{R}}_{i,k}^\top&\tilde{b}_{i,k}\end{bmatrix}^\top\\
\boldsymbol{\omega}_{0,k}& = \begin{bmatrix}\boldsymbol{\omega}_{\mathbf{p}_k}^\top&\boldsymbol{\omega}_{\mathbf{R}_k}^\top\end{bmatrix}^\top,\boldsymbol{\omega}_{i,k}= \begin{bmatrix}\omega_{b_{i,k}}&\omega_{\theta_{i,k}}\end{bmatrix}^\top
\end{align}
where $\boldsymbol{\omega}_{0,k}$ and $\boldsymbol{\omega}_{i,k}$ represent the process noise vectors.

On the basis of (1)-(6), the estimated values obtained from the model's prediction are as follows ($i>0$):
\begin{align}
\mathbf{\hat{p}}_{0,k}^0 &= \mathbf{\bar{p}}_{0,k-1}, \quad\mathbf{\hat{R}}^0_{0,k} = \mathbf{\bar{R}}_{0,k-1}\label{root prediction}\\
\mathbf{\hat{p}}_{i,k}^0 &= \mathbf{\bar{p}}_{i-1,k} + \mathbf{\bar{R}}_{i-1,k}{}^{i-1}\!\mathbf{p}_i\\
\mathbf{\hat{R}}_{i,k}^0 &= \mathbf{\bar{R}}_{i-1,k}\mathrm{Exp}\left(\boldsymbol{\theta}_{i,k}-{b}_{i,k-1}\frac{\boldsymbol{\theta}_{i,k}}{\|\boldsymbol{\theta}_{i,k}\|}\right)\label{bias prediction}\\
{\hat{b}}_{i,k}^0 &={\bar{b}}_{i,k-1}
\end{align}
Using (\ref{error state vec}) and (\ref{error state so3}), the true values on the left-hand sides of (1), (2), and (4)-(6) can be expressed in terms of the error state $\mathbf{\tilde{x}}$ relative to the predicted value and the predicted value itself $\mathbf{\hat{x}}^0$. 
Similarly, the true values on the right-hand side can be expressed via the error state $\mathbf{\tilde{x}}$ relative to the estimated value and the estimated value itself $\mathbf{\bar{x}}$. 
These allow us to rewrite the articulated structure model as an equation for the transition of the error state.
The derived equation can be linearized in the area where the error and noise approach zero and can be expressed as follows ($i>0$):
\begin{align}
\mathbf{\tilde{x}}_{0,k}
&\simeq\mathbf{F}_{\mathbf{\tilde{x}}_{0,k-1}}\mathbf{\tilde{x}}_{0,k-1}+\mathbf{F}_{\boldsymbol{\omega}_{0,k}}\boldsymbol{\omega}_{0,k}\label{linearize root}\\
\mathbf{\tilde{x}}_{i,k}
&\simeq\mathbf{F}_{\mathbf{\tilde{x}}_{i,k-1}}\mathbf{\tilde{x}}_{i,k-1}+\mathbf{F}_{\mathbf{\tilde{x}}_{i-1,k}}\mathbf{\tilde{x}}_{i-1,k}+\mathbf{F}_{\boldsymbol{\omega}_{i,k}}\boldsymbol{\omega}_{i,k}\label{linearize link}
\end{align}
where each $\mathbf{F}$ is a Jacobian for linearization and can be derived as in \cite{fastlio}.
Since the state variables propagated in the spatial direction, $\mathbf{x}_{i-1,k}$, and those propagated in the time direction, $\mathbf{x}_{i,k-1}$, do not influence each other, $\mathbf{\tilde{x}}_{i-1,k}$ and $\mathbf{\tilde{x}}_{i,k-1}$ can be expressed as sums of mutually independent vectors as in (\ref{linearize link}).
With (\ref{linearize root}) and (\ref{linearize link}), the uncertainty of the error state is propagated as follows ($i>0$):
\begin{align}
\mathbf{\hat{P}}_{0,k} &= \mathbf{F}_{\mathbf{\tilde{x}}_{0,k-1}}\mathbf{\bar{P}}_{0,k-1} \mathbf{F}_{\mathbf{\tilde{x}}_{0,k-1}}^\top+\mathbf{F}_{\boldsymbol{\omega}_{0,k}}\mathbf{Q}_{0,k}\mathbf{F}_{\boldsymbol{\omega}_{0,k}}^\top \label{head propagation}\\
\begin{split}
\mathbf{\hat{P}}_{i,k} &= \mathbf{F}_{\mathbf{\tilde{x}}_{i,k-1}}\mathbf{\bar{P}}_{i,k-1}\mathbf{F}_{\mathbf{\bar{x}}_{i,k-1}}^\top+ \mathbf{F}_{\mathbf{\tilde{x}}_{i-1,k}}\mathbf{\bar{P}}_{i-1,k}\mathbf{F}_{\mathbf{\tilde{x}}_{i-1,k}}^\top\\
&+\mathbf{F}_{\boldsymbol{\omega}_{i,k}}\mathbf{Q}_{i,k}\mathbf{F}^\top_{\boldsymbol{\omega}_{i,k}}\label{link propagation}
\end{split}
\end{align}
where $\mathbf{\hat{P}}_{0,k}$ and $\mathbf{\hat{P}}_{i,k}$ are propagated covariance matrices of $\mathbf{\tilde{x}}_{0,k}$ and $\mathbf{\tilde{x}}_{i,k}$; $\mathbf{\bar{P}}_{0,k-1}$, $\mathbf{\bar{P}}_{i,k-1}$ and $\mathbf{\bar{P}}_{i-1,k}$ are covariance matrices of $\mathbf{\tilde{x}}_{0,k-1}$, $\mathbf{\tilde{x}}_{i,k-1}$ and $\mathbf{\tilde{x}}_{i-1,k}$, respectively; and $\mathbf{Q}_{0,k}$ and $\mathbf{Q}_{i,k}$ are the noise covariances of $\boldsymbol{\omega}_{0,k}$ and $\boldsymbol{\omega}_{i,k}$, which are set manually.

As with the articulated structure model, the measurement model in (\ref{measurement model2}) can be rewritten using the error state and can be linearized as follows:
\begin{align}
0&\simeq{z}_j^\kappa+\mathbf{H}_j^\kappa\mathbf{\tilde{x}}_{i,k}^\kappa+{v}_j\label{linearized measurement model}
\end{align}
where ${z}_j^\kappa$ represents the actual observed measurement obtained by substituting $\mathbf{p}_{i,k}=\mathbf{\hat{p}}_{i,k}^\kappa$, $\mathbf{R}_{i,k}=\mathbf{\hat{R}}_{i,k}^\kappa$ and $\mathbf{v}_j=\mathbf{0}$ into (\ref{measurement model2}) and serves as the basis for linearization.
$\mathbf{H}_j^\kappa$ is a Jacobian for linearization, corresponding to $\mathbf{\tilde{x}}_{i,k}^\kappa$. ${v}_j = -\mathbf{n}_j\mathbf{\hat{R}}_{i,k}^\kappa\mathbf{R}_{S_i}^{L_i}\boldsymbol{\upsilon}^j_{i,k}$ includes measurement noise whose covariance varies with each estimation, but it is shown in \cite{fastlio2} that setting this variable to a constant value works well.
As $\kappa$ is included in (\ref{linearized measurement model}), the observation model is computed for each iteration.

The iterated Kalman filter estimates the increment $\mathbf{\tilde{x}}^\kappa$ with respect to the current error state vector $\delta\mathbf{\hat{x}}^\kappa$ to minimize the following weight square sum:
\begin{align}
\underset{\mathbf{\tilde{x}}^\kappa}{\mathrm{min}}\left(\|\mathbf{\tilde{x}}^{0,\kappa}_{i,k}+\mathbf{J}^\kappa_{i,k}\mathbf{\tilde{x}}^\kappa_{i,k}\|+\sum_{j=1}^m\|{z}^\kappa_j+\mathbf{H}^\kappa_j\mathbf{\tilde{x}}_{i,k}^\kappa\|\right)\label{weight square sum}
\end{align}
where $\mathbf{J}^\kappa_{i,k}$ is a square matrix that eliminates the nonlinearity associated with the computation of the attitude error vector \cite{fastlio2,fastlio} and where $m$ is the number of point measurements.

With (\ref{weight square sum}), we can estimate $\mathbf{\tilde{x}}_{i,k}^\kappa$ as follows \cite{fastlio2,fastlio}:
\begin{align}
\mathbf{z}^\kappa& = \begin{bmatrix}z_1^\kappa&\ldots&z^\kappa_m\end{bmatrix},\quad \mathbf{H}^\kappa = \begin{bmatrix}{\mathbf{H}_{1}^\kappa}^\top&\ldots&{\mathbf{H}^\kappa_m}^\top\end{bmatrix}^\top \label{H and Z}\\
\mathbf{K}_{i,k}^\kappa& = \left(\mathbf{H}^\top_\kappa\mathbf{R}^{-1}\mathbf{H}_\kappa+\left(\mathbf{J}^\kappa_{i,k}\right)^\top\mathbf{\hat{P}}^{-1}_{i,k}\mathbf{J}^\kappa_{i,k}\right)^{-1}\mathbf{H}^\top_\kappa\mathbf{R}^{-1} \label{Kalman gain}\\
\mathbf{\tilde{x}}_{i,k}^{\kappa}&=\mathbf{K}^\kappa_{i,k}\left(-\mathbf{z}^\kappa_{i,k}+\mathbf{H}_{i,k}^\kappa\left(\mathbf{J}^\kappa_{i,k}\right)^{-1}\mathbf{\hat{x}}_{i,k}\right) -\left(\mathbf{J}^\kappa_{i,k}\right)^{-1}\mathbf{\hat{x}}_{i,k}^\kappa \label{update redisual}
\end{align}
where $\mathbf{R}$ represents the diagonal covariance matrix of $\upsilon_1$ to $\upsilon_j$.
In (\ref{Kalman gain}), we obtain a variant of the formula for the general Kalman gain $\mathbf{K}_{i,k}^\kappa$ by using an inverse matrix lemma, allowing the calculation to be performed in the state dimension rather than the measurement dimension \cite{fastlio2}.
From the estimate $\mathbf{\tilde{x}}^\kappa_{i,k}$, each state is updated as follows.
\begin{align}
\mathbf{\hat{p}}_{i,k}^{\kappa+1} &= \mathbf{\hat{p}}^\kappa_{i,k} +\mathbf{\tilde{p}}^\kappa_{i,k}\\
\mathbf{\hat{R}}^{\kappa+1}_{i,k} &= \mathbf{\hat{R}}^\kappa_{i,k}\mathrm{Exp}\left(\mathbf{\tilde{R}}^\kappa_{i,k}\right)\\
\hat{b}^{\kappa+1}_{i,k} &= \hat{b}^\kappa_{i,k} +\tilde{b}^\kappa_{i,k} \label{bias update}
\end{align}
Using $\mathbf{\hat{x}}^{\kappa+1}$, (\ref{linearized measurement model})-(\ref{bias update}) are repeated until $\mathbf{\tilde{x}}^\kappa_{i,k}$ falls below the threshold and converges.
After convergence, the final estimated state $\mathbf{\bar{x}}_{i,k}$ and the covariance matrix of its error state $\mathbf{\bar{P}}_{i,k}$ are determined as follows:
\begin{align}
\mathbf{\bar{x}}_{i,k} &= \mathbf{\hat{x}}^{\kappa+1}_{i,k} \label{state update}\\
\mathbf{\bar{P}}_{i,k}&= \left(\mathbf{I}-\mathbf{K}^\kappa\mathbf{H}^\kappa\right)\left(\mathbf{J}^\kappa\right)^{-1}\mathbf{\hat{P}}_{i,k}\left(\mathbf{J}^\kappa\right)^{-\top} \label{covariance update}
\end{align}   
The estimated state and covariance are propagated in the spatial and temporal directions and are used for each estimation.
The points acquired by the proximity sensor $\mathbf{p}^j_{i,k}$ are transformed into the world frame ${}^G\!\mathbf{p}^j_{i,k}$ on the basis of the estimated state of the link and added to the map by {\it on-tree downsampling} \cite{fastlio2} after each link estimation step.
\begin{align}
{}^G\!\mathbf{p}_{i,k}^j=\mathbf{\bar{R}}_{i,k}\left(\mathbf{R}_{S_i}^{L_i}\mathbf{p}^j_{i,k}+\mathbf{p}_{S_i}^{L_i}\right)+\mathbf{\bar{p}}_{i,k} \label{map update}
\end{align}
In summary, the estimation process of this method is shown in Algorithm 1.
\begin{algorithm}[tb]
\caption{Estimation process}
\For{$k$}
{
  \For {$0\le i \le $ \rm{number of links}}
  {
	\If {$i=0$}{
		{\bf Input:} $\mathbf{\bar{x}}_{0,k-1}$, $\mathbf{\bar{P}}_{0,k-1}$, $\mathbf{p}_j$\;
		Calculate $\mathbf{\hat{P}}_{0,k}$ by (\ref {head propagation})\;
	}
	\Else{
		{\bf Input:} $\mathbf{\bar{x}}_{i,k-1}$, $\mathbf{\bar{P}}_{i,k-1}$, $\mathbf{\bar{x}}_{i-1,k}$, $\mathbf{\bar{P}}_{i-1,k}$\;
		{\bf Input:} $\mathbf{p}_j$, and $\theta_{i,k}$\;
		Calculate $\mathbf{\hat{x}}^0_{i,k}$, $\mathbf{\hat{P}}_{i,k}$ by (\ref {link propagation})\;
	}
	$\kappa \gets 0$\;
	\Repeat {$\mathbf{\tilde{x}}^\kappa_{i,k}$ \rm{converges}}{
		Compute $\mathbf{H}^\kappa$, $\mathbf{z}^\kappa$, $\mathbf{J}^\kappa$, and $\mathbf{\tilde{x}}^{0,\kappa}$\;
		Compute $\mathbf{\tilde{x}}^{\kappa}$ and $\mathbf{\hat{x}}^{\kappa+1}$ via (\ref{Kalman gain})-(\ref{bias update})\;
		$\kappa\gets\kappa+1$\;
	}
	{\bf Output:} $\mathbf{\bar{x}}_{i,k}$, $\mathbf{\bar{P}}_{i,k}$ via (\ref{state update}), (\ref{covariance update})\;
	Add the measured points to the map according to $\mathbf{\bar{x}}_{i,k}$ via (\ref{map update})\;
  }
}
\end{algorithm}

\section{Simulation}\label{Simulation}
To verify the effectiveness of the proposed method, we conducted simulations in Gazebo under multiple environments. 
The multijoint structure model in the simulation consists of links connected in series by one-DOF joints.
In Fig.~\ref{system-overview}, we define the x-axis along the length of the links, the y-axis in the depth direction, and the z-axis in the vertical direction. 
For this simulation, we assume a model in which odd-numbered joints rotate around the z-axis and even-numbered joints rotate around the y-axis.
Eight proximity sensors, modeled after the VL53L5C, are positioned circumferentially at the midpoint of each link's length with a radius of 5 cm.
Each sensor acquires the data of 64 points, resulting in a total of 512 points per link.
The detection range of each sensor is 5 cm to 4 m.
Furthermore, we consider two sources of uncertainty: a bias of 0.05 rad in the angle measurement for each joint and white noise in the distance measurements obtained from proximity sensors.
The covariance $\sigma$ of the white noise is set to $2.7\sigma = 0.01r$, where $r$ is the measured distance.

As described in Section \ref{system overview}, the proposed method is divided into the steps of estimating the root and estimating each link following the root. 
Since the method of root estimation is the same as in the general SLAM method, it is not the primary focus of this paper.
The most crucial point of our proposed method is the estimation of each link following the root, which combines state variables that propagate spatially and temporally.
Therefore, in Section \ref{fixed simulation}, we first present the results of simulations in which the root is fixed.
This allows us to verify whether the state of the entire structure relative to the environment can be accurately determined when the root state with respect to the environment is known.
Additionally, in Section \ref{free simulation}, to verify the adaptability of the proposed method to situations where the root position is ambiguous or the root is in motion, we conduct simulations with an unconstrained root.

\subsection{Effects of Spatial Direction Estimation}\label{fixed simulation}
We evaluated the proposed method via two types of structures with fixed roots: a 5-link structure assumed to represent a robot arm and a 20-link structure assumed to represent a robot with many 
small-scale
 links, such as a snake robot.
The link lengths of the articulated structures were 40 cm for the 5-link structure and 17 cm for the 20-link structure.
This allowed us to investigate the performance differences due to the variation in the number of links.
To generate motion, arbitrary sine waves were given as position command values to each joint of the simulation model.
We then compared the state of the articulated structure relative to the environment, which was obtained by the proposed method, with that derived solely from the kinematic model to verify the effectiveness of the proposed method. 
\begin{figure*}[t]
\begin{center}
\includegraphics[width=2\columnwidth]{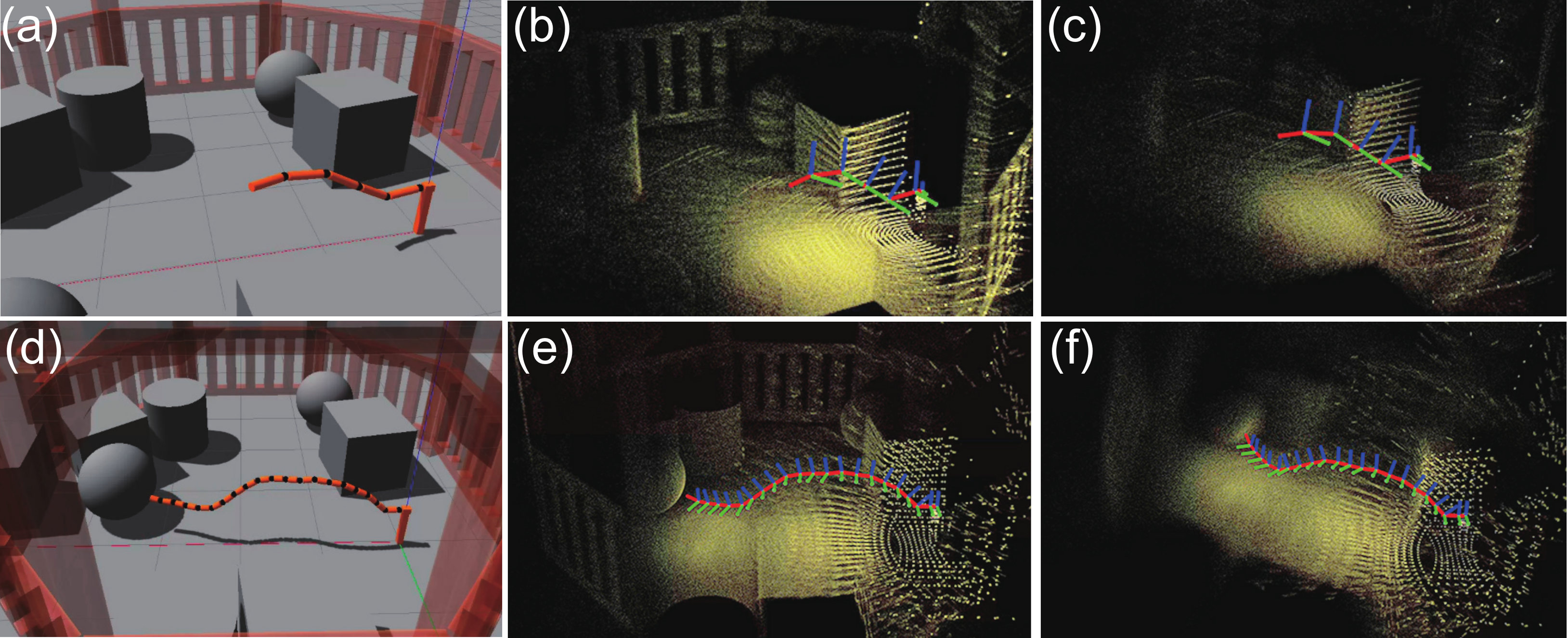}
\caption{A snapshot of the simulation. (a), (d) show the actual structure in the simulation environment; (b), (e) show the states and acquisition environment obtained via the proposed method; and (c), (f) show the states and acquisition environment obtained via the kinematic model.
The position and orientation of the robot are represented in coordinate systems, where the origin of each system is located at the center of the adjacent joint on the root side of each link. In these coordinate systems, the x-axis is denoted by red, the y-axis by green, and the z-axis by blue.}
\label{snapshot fixed}
\end{center}
\end{figure*}
Fig.~\ref{snapshot fixed} (a)-(c) present the results of the simulations with the 5-link structure, whereas (d)-(f) present those for the 20-link structure. 
Specifically, (a) and (d) show the configuration of the structure in the simulation, (b) and (e) present the structure's state and the mapped environment obtained by the proposed method, and (c) and (f) show the structure state obtained by the kinematic model and the acquisition point cloud projected on the basis of it.
(b) and (c), as well as (e) and (f), are captured with the same viewing angle relative to the root.

In the case of the 5-link structure, the kinematic model resulted in an average deviation exceeding 12.5 cm from the true end-effector position in a 100-second simulation due to accumulated bias of each link.
Consequently, misalignment occurred in the surrounding environment data acquired by each sensor, leading to an indistinct map.
In contrast, the proposed method successfully constrained the end-effector error to within 5 cm, reducing the average positional error in a 100-second simulation by approximately 62\% while obtaining a clear map of the sensing environment.
\begin{figure}[tb]
\begin{center}
\includegraphics[width=\columnwidth]{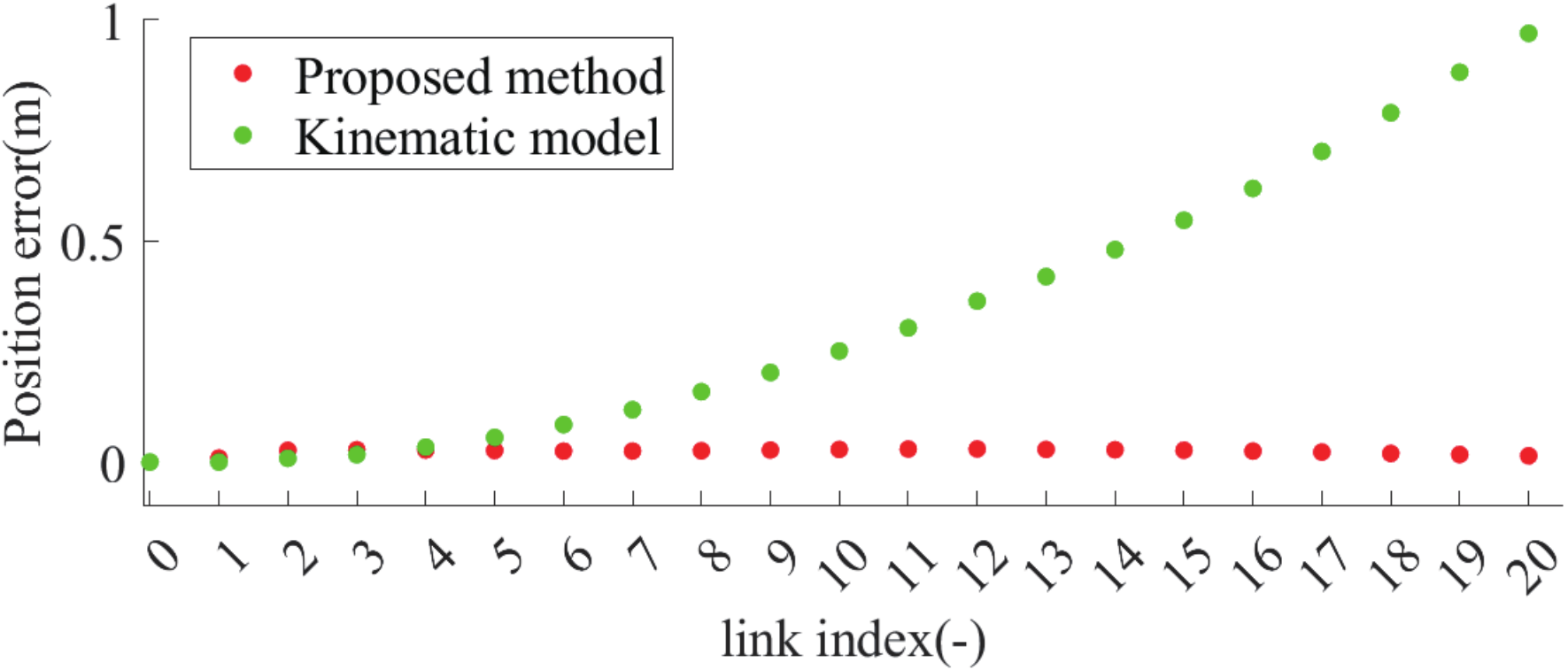}
\caption{Average absolute value of the error with respect to the true value of the position for each link.}
\label{average}
\end{center}
\end{figure}

\begin{figure}[tb]
\begin{center}
\includegraphics[width=\columnwidth]{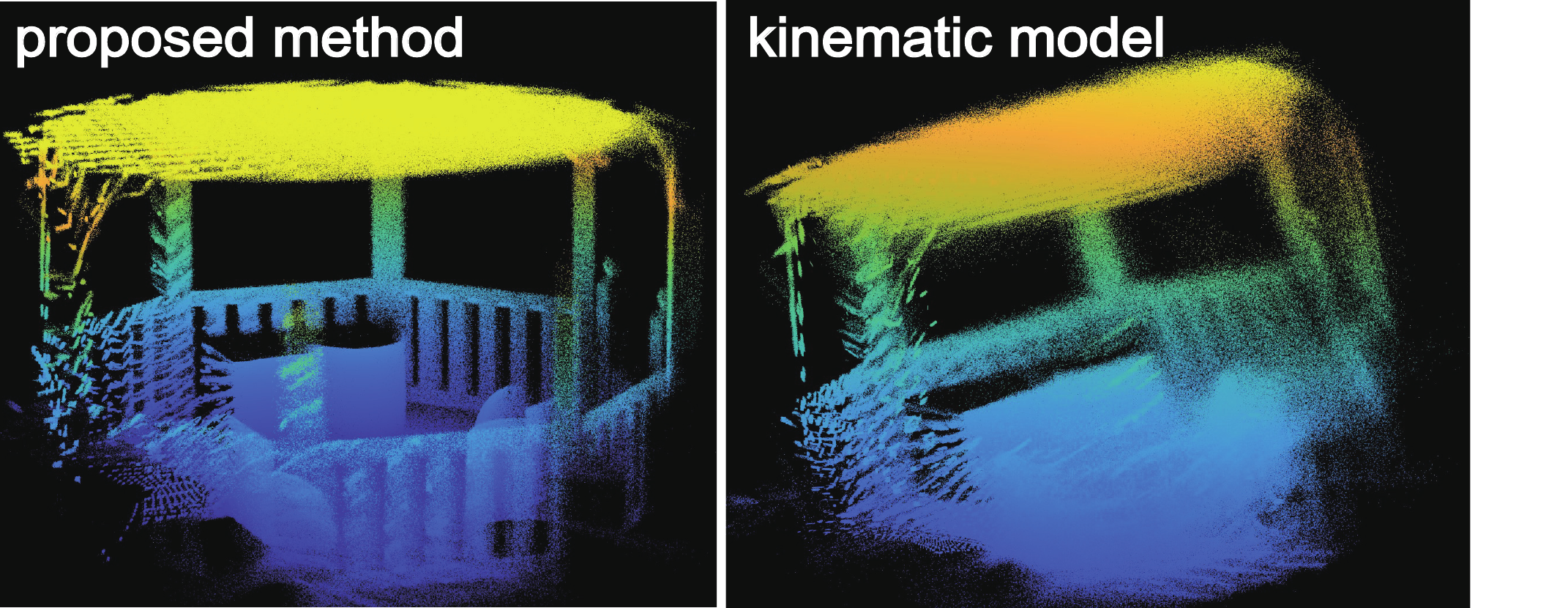}
\caption{Surrounding environment acquired from the entire structure.}
\label{mapping}
\end{center}
\end{figure}

As the number of links increases, the cumulative bias also increases. 
Consequently, the state of the 20-link structure obtained from the kinematic model deviates significantly from the actual state, as illustrated in Fig.~\ref{snapshot fixed}(f).
As illustrated in (e), the proposed method was able to accurately estimate the full-body state relative to the environment, even in the presence of significant errors caused by cumulative inaccuracies.
Fig.~\ref{average} presents the plot of the average absolute errors for each link in the 20-link simulation over 100 seconds. 
Owing to the accumulation of bias, the estimation based solely on the kinematic model shows an increase in error with each successive link. 
In contrast, the proposed method demonstrates the ability to estimate the position of each link within a 5 cm error, regardless of the number of links.
Furthermore, Fig.~\ref{mapping} depicts the external environment map obtained during this process, demonstrating that the proposed method significantly improves the acquired environmental information in line with the enhanced state estimation accuracy.

Through simulations with a fixed root, the results demonstrate the effectiveness of spatial propagation in estimation, which is the most crucial aspect of our proposed method. 
A detailed discussion of this topic is provided in Section \ref{Advantages of Estimation in Spacial Direction}. Furthermore, by conducting simulations on articulated structures with varying numbers of links, we show that the proposed method performs well regardless of the number of links and the extent of cumulative error.

\subsection{Robustness to Root Uncertainty}\label{free simulation}
\begin{figure}[tb]
\begin{center}
\includegraphics[width=\columnwidth]{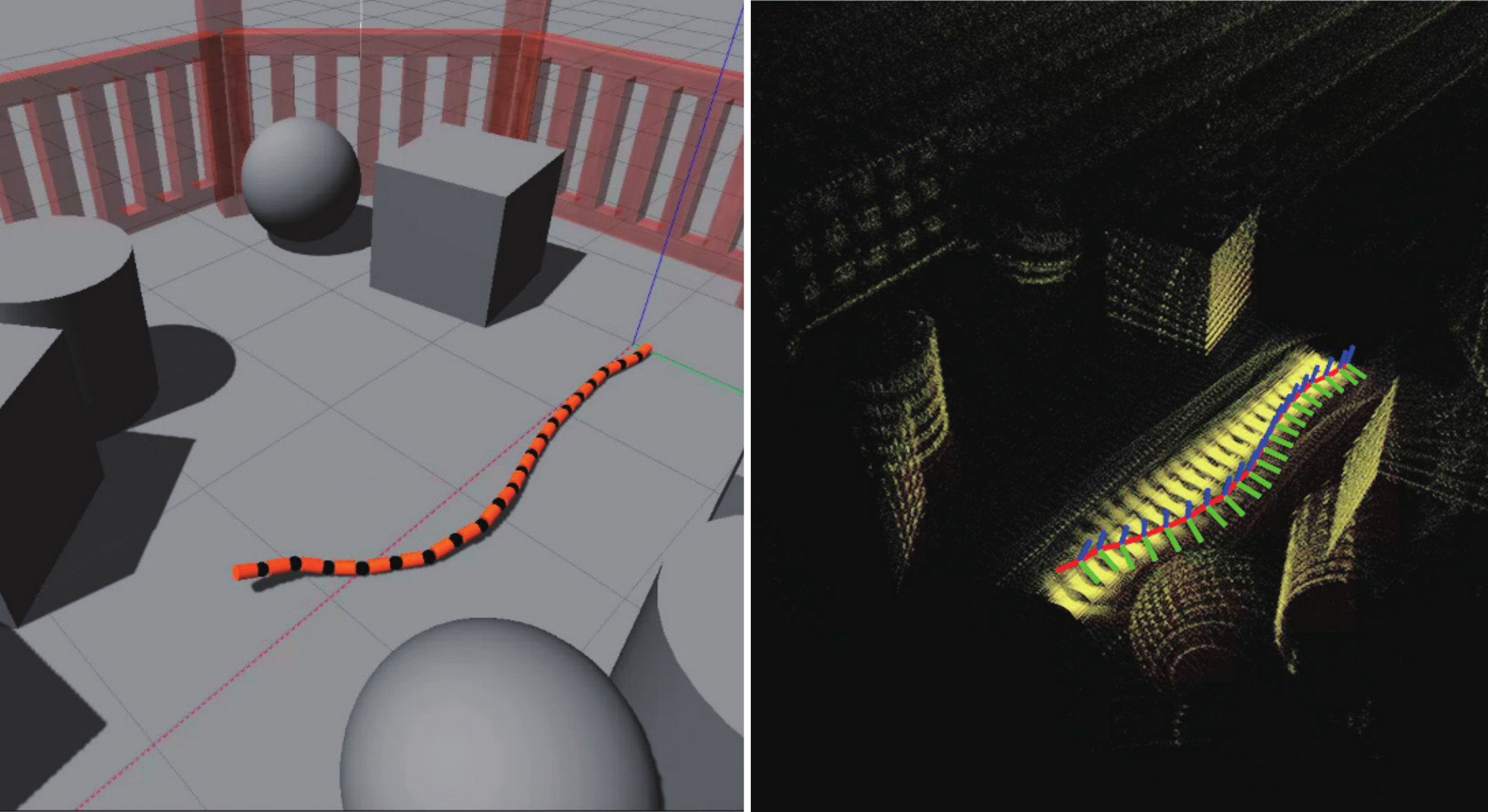}
\caption{Simulation with an unconstrained root. The left image shows the actual state of the articulated structure in the simulation, whereas the right image displays the estimated state of the structure and the acquired surrounding environment.}
\label{snapshot free}
\end{center}
\end{figure}

\begin{figure}[tb]
\begin{center}
\includegraphics[width=\columnwidth]{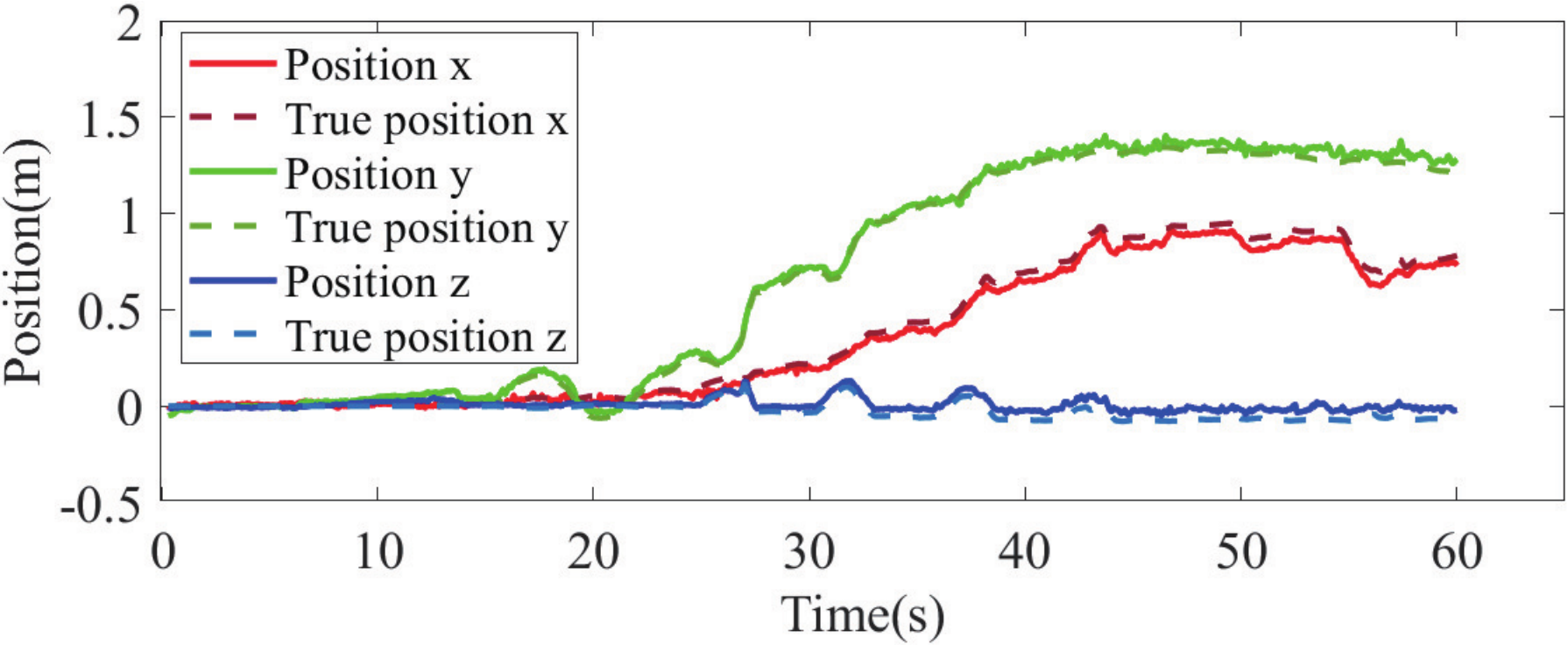}
\includegraphics[width=\columnwidth]{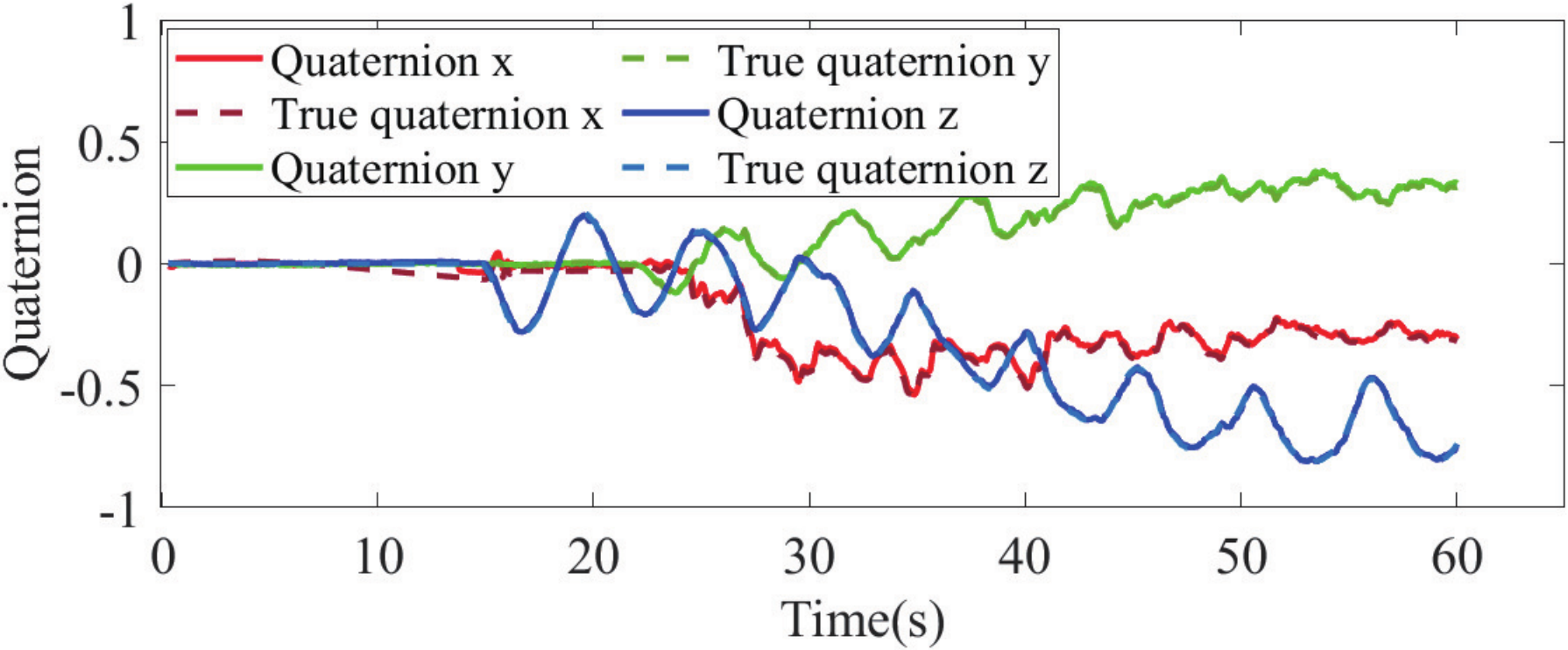}
\caption{Plot of the values estimated via the proposed method and the ground truth for the root. The upper plot shows the position, and the lower plot displays the x, y, and z components of the quaternion.}
\label{free plot}
\end{center}
\end{figure}

As shown in Fig.~\ref{snapshot free}, we placed the 20-link articulated structure on the ground, similar to the simulation in Section \ref{fixed simulation}, but did not fix it in place. 
The motion of the structure is generated from a random sine wave applied to each joint, and no control is implemented.
The biases in each joint and the noise of the proximity sensors were maintained at the same levels as those in the fixed-root simulation in Section \ref{fixed simulation}.
In this study, the IMU is not used to estimate the root, and the model prediction equation is shown in \eqref{root prediction}.
Hence, it is impossible to understand the movement of the root in relation to the environment without environmental information, and a comparison with the results obtained using the kinematic model alone is not given here.
Fig.~\ref{free plot} depicts the trajectory of the root's position and orientation, along with the corresponding ground-truth trajectory.
The orientation is represented via quaternions, and the $x$, $y$, and $z$ components of the quaternions are plotted in Fig.~\ref{free plot}.
From 15 s onward, the position and orientation of the root undergo displacement due to the whole-body motion of the structure. 
Despite this movement, our method successfully estimates the orientation of the root on the basis of the environmental information obtained from proximity sensors distributed over the whole body.
However, the estimation is unsuccessful in cases where the root undergoes rapid movements, such as when the entire structure collapses in a rolling motion.
This failure can be attributed to the relatively low temporal frequency of sensor data acquisition compared with the speed of movement, resulting in temporally sparse information.

In simulations with unfixed roots, we were able to demonstrate that the proposed method can be effectively applied even when the entire structure moves relative to the environment and the state of the root is unclear, provided that the movement of the root is slow.

\section{Discussion}\label{Discussion}
\subsection{Advantages of Estimation in the Spatial Direction}\label{Advantages of Estimation in Spacial Direction}
Distributing proximity sensors over an entire structure not only enables the acquisition of environmental information surrounding the entire structure but also allows the individual estimation of each link by providing unique external information to each joint. 
This estimation process enables the spatial propagation of the estimated state quantities.
In the simulation of the 20-link articulated structure described in Section \ref{fixed simulation}, we considered a scenario where, owing to the assumed biases in each link, the end-point position obtained from the kinematic model exhibited an error magnitude of up to 1 m relative to the total length of 3 m of the structure.
The ability of the proposed method to accurately estimate the states in such scenarios can be attributed to its approach of propagating the estimation in the spatial direction and correcting errors from the root. 
Furthermore, as the estimation of each link is followed by the sequential incorporation of the acquired environmental information into the map, the increased availability of environmental data to the end-effector links contributes significantly to improving the estimation accuracy of these terminal links.

The benefits of propagating this spatial information are due to the fact that all links are estimated simultaneously.
Therefore, when designing the hardware system, the ideal design would be one in which all the proximity sensors across the body can acquire data as close to simultaneously as possible.
Furthermore, as a future development of the proposed method, we are considering the possibility of developing a technique that can accommodate temporal misalignment of sensor data acquisition in cases where simultaneous data collection is difficult.

\subsection{Contribution to the Estimation of Roots}
In Section \ref{free simulation}, we verified that when the movement per unit time is small, it is possible to estimate the root's motion using only external environmental information, even if the structure is not fixed relative to the environment.
The map utilized for estimation in the proposed method incorporates environmental information acquired by proximity sensors on links other than the one being estimated. 
This comprehensive approach enables the estimation process to leverage all available environmental data.
The utilization of a map covering a broader area than that captured by the root's own sensors is likely a contributing factor to successful estimation, even in the absence of precise temporal information about the root's movement.

\subsection{Application}
Our proposed method can adapt to more various situations by obtaining higher-frequency information about the root's state.
In Section \ref{free simulation}, the estimation was performed under the assumption that the temporal changes in the root were unknown, which made it difficult to estimate rapid movements. 
However, if we can obtain the state between sampling periods of proximity sensors via devices such as IMUs or wheel odometry, we can improve the accuracy of the estimation.
By incorporating information from these sensors into the model equations (\ref{root pos model}) and (\ref{root rot model}), we can expect to achieve whole-body state estimation relative to the environment and environmental information acquisition for articulated mobile robots involving large displacements and rotations.

Furthermore, although this paper considers a basic model in which joint angles are directly obtained from encoders, we believe that our method can be applied to any articulated structure or soft robot, regardless of the input, as long as the structure can be spatially discretized and the relative relationships between each link can be described by a geometric model.
In future work, we aim to implement this method on various articulated structures or soft robots to verify its effectiveness.

\section{CONCLUSION}\label{Conclusion}
We propose a method for whole-body state estimation and environmental information acquisition for articulated structures that are challenging to assess via conventional kinematic models due to link deformations or that require comprehensive external environmental information. 
This method involves deploying proximity sensors throughout the body.
The proximity sensors distributed across the entire structure not only enable the acquisition of environmental information surrounding the whole structure but also facilitate individual joint estimation, allowing estimation to progress spatially.
This spatial progression of estimation enables the correction of cumulative biases along the length of the structure from the root, thus allowing for accurate whole-body state estimation even when the state significantly deviates from the kinematic model.
Our method was validated through simulations, which demonstrated its ability to correct individual joint biases in a model with inherent biases and to achieve accurate posture estimation relative to the actual environment.
The proposed method is intended to be applied to various articulated structures and soft robots, and we would like to implement it on nonrigid robots or articulated mobile robots to verify its effectiveness in future work.

\addtolength{\textheight}{-12cm}   




\section*{ACKNOWLEDGMENT}

This work was partially supported by JSPS KAKENHI,
Grant Numbers JP23K20923 and JP24H00726.


\bibliographystyle{IEEEtran} 
\bibliography{IEEEabrv,test_iwao}

\end{document}